\documentclass[lettersize,journal]{IEEEtran}
\pdfoutput=1 
\usepackage{amsmath,amsfonts}
\usepackage{algorithmic}
\usepackage{algorithm}
\usepackage[export]{adjustbox}
\usepackage{array}
\usepackage[caption=false,font=normalsize,labelfont=sf,textfont=sf]{subfig}
\usepackage{textcomp}
\usepackage{stfloats}
\usepackage{url}
\usepackage{verbatim}
\usepackage{graphicx}
\usepackage{cite}
\usepackage[hidelinks]{hyperref}
\graphicspath{ {./figs/} }
\usepackage{textgreek}
\usepackage{float}
\usepackage{mathtools}
\usepackage{color}
\usepackage{ragged2e}
\usepackage{longtable}
\usepackage{csquotes}
\usepackage{lipsum,multicol}

\usepackage{tabularx,booktabs}
\newcolumntype{C}{>{\centering\arraybackslash}X} 
\usepackage{lipsum} 

\begin{document}

\title{A Bayesian Optimization approach for calibrating large-scale activity-based transport models}

\author{Serio Agriesti$^{1,2}$, Vladimir Kuzmanovski$^{2,3,4}$, Jaakko Hollmén$^{3,5}$,~\IEEEmembership{Senior Member,~IEEE}, Claudio Roncoli$^{1}$, and Bat-hen Nahmias-Biran$^{6}$
\thanks{$^{1}$Department of Built Environment, School of Engineering, Aalto University, Espoo 02150 Finland
{\tt\small serio.agriesti@aalto.fi, claudio.roncoli@aalto.fi}}%
\thanks{$^{2}$
FinEst Centre for Smart Cities, Tallinn University of Technology, Tallinn, Estonia}%
\thanks{$^{3}$Department of Computer Science, Aalto University, Espoo 02150 Finland
{\tt\small vladimir.kuzmanovski@aalto.fi}}%
\thanks{$^{4}$
Department of Knowledge Technologies, Joˇzef Stefan Institute, Slovenia}%
\thanks{$^{5}$Department of Computer and Systems Sciences, Stockholm University, Sweden
{\tt\small jaakko.hollmen@dsv.su.se}}%
\thanks{$^{6}$
Department of Civil Engineering, Ariel University, Ramat HaGolan, Ariel 40700, Israel
{\tt\small bathennb@ariel.ac.il}}%
\thanks{This research is partly funded by the FINEST Twins Center of Excellence, H2020 European Union funding for Research and Innovation grant number 856602, and by the Academy of Finland project ALCOSTO (349327).}
\thanks{Manuscript received --; revised --}}



\maketitle

\begin{abstract}
The use of Agent-Based and Activity-Based modeling in transportation is rising due to the capability of addressing complex applications such as disruptive trends (e.g., remote working and automation) or the design and assessment of disaggregated management strategies.
Still, the broad adoption of large-scale disaggregate  models is not materializing due to the inherently high complexity and computational needs. Activity-based models focused on behavioral theory, for example, may involve hundreds of parameters that need to be calibrated to match the detailed socio-economical characteristics of the population for any case study. 
This paper tackles this issue by proposing a novel Bayesian Optimization approach incorporating a surrogate model in the form of an improved Random Forest, designed to automate the calibration process of the behavioral parameters.
The proposed method is tested on a case study for the city of Tallinn, Estonia, where the model to be calibrated consists of 477 behavioral parameters, using the SimMobility MT software. 
Satisfactory performance is achieved in the major indicators defined for the calibration process: the error for the overall number of trips is equal to 4\% and the average error in the OD matrix is 15.92 vehicles per day.
\end{abstract}

\begin{IEEEkeywords}
Activity-based transport modeling, Model calibration, Machine learning, Bayesian optimization, Surrogate model.
\end{IEEEkeywords}

\section{Introduction}
\IEEEPARstart{L}{arge-scale} transportation problems have always been prone to high complexity, approximations, and a lack of a univocal mathematical formulation \cite{TRR1}. This is especially the case for models involving human behavior and the numerous factors ruling over mobility choices, which have been applied to narrow scopes (e.g.,~modal choices) or have been designed as aggregated (e.g., ~four-step models). Still, current and future transportation challenges -- e.g. urbanization, population growth, and congestion \cite{UN:report}, \cite{UMR2019} but also disruptive events such as pandemics \cite{UMR2021} or the climate crisis -- require solutions able to frame transport demand through the lenses of individual choices on a large scale. 
Future innovations on the transport supply spectrum \cite{FT-CoE} are also foreseen to have disruptive effects on mobility demand. To evaluate said innovations, tools able to frame changed mobility choices and travel habits are needed. Currently, agent-based modeling (ABM) is the most promising solution due to the ability to frame both demand and supply at the agent level (the agent being either the individual or the single vehicle) in a disaggregate fashion, allowing the investigation of emergent behaviors \cite{ABMLit}. Specifically, activity-based models are a particular kind of ABM that considers each individual in the population as an agent, while modeling their behavioral choices in a disaggregated fashion. Activity-based and ABM models have already successfully been exploited to carry out policy analyses \cite{TRR3}, accessibility studies \cite{Logsum}, and forecasting experiments \cite{SingapSimMob, TRR2, MunichMatsim} involving automated mobility. Automated vehicles are a perfect example of the potentiality of a disaggregate approach, in which aspects such as the fleet size and the routing algorithm need to be assessed based on a realistic demand, rooted in behavioral models rather than in historical patterns. Another ABM application addressing the reported long-term challenges and strongly benefitting from a properly calibrated large-scale activity-based model is the modeling of remote and hybrid working patterns during public health crises \cite{TuDelftThesis}.

\IEEEpubidadjcol
Still, despite these needs, the state-of-the-art concerning activity-based models behind most ABM is limited, especially when it comes to the underlying calibration. The different structures of the available activity-based tools and the different magnitude of parameters to be calibrated each time are still a barrier against a wider adoption. A unified calibration approach, not dependent on specific software and able to include hundreds of parameters, is still needed to fill this research gap and foster the usage of behavioral activity-based models \cite{ActBMCalib-recent, AmericanRegionalHandbook}. This paper tries to accomplish the above, enabling wider adoption of activity-based models by proposing a method for calibration via efficient global optimization -- the Bayesian optimization (BO). BO \cite{shahriari2015taking, gutmann2016bayesian} seeks the global optimum through a sequential sampling design and approximation of an underlying likelihood function using a surrogate model and surrogate (response) surface. The evaluation of the surrogate model requires significantly less computational time, making it suitable for extensive utilization in guiding the search process and balancing the exploration-exploitation trade-off. As such, the method is designed for optimization problems that feature \enquote{expensive} functions in terms of computational time, which are approximated through the surrogate surface.
This study focuses on the development of a high-dimensional BO method that converges within a given computational budget and avoids the necessity to master the implementation of the underlying large-scale ABM. The proposed algorithm and the modifications to the BO approach are designed to be transferable to other large-scale problems involving dozens of parameters and fit to be analyzed through a surrogate model. 

\indent In Section~\ref{Literature review}, a review of the current literature is carried out, concerning the calibration of large-scale transport models and BO; Section III elaborates on the proposed methodology; Section IV introduces the case study, while Section V reports the results obtained by applying the proposed methodology to the case study; Section VI discusses the results and highlights the main conclusions for this work.

\section{Literature review} \label{Literature review}
\subsection{Calibration approaches in transport modeling}
The calibration of activity-based models in transport has received far less attention than the calibration of other ABMs, while existing approaches mainly rely on heuristics, which, in turn, is hindering the potentialities of these tools.

The work presented in \cite{ActBMCalib-recent} describes a gradient- and simulation-based optimization procedure designed to calibrate 28 parameters in a utility-based nested logit system. Similarly, \cite{LiThesis} exploits the WSPSA algorithm to calibrate 94 behavioral parameters on the demand side, still, it does not exploit a surrogate model and thus needs multiple computationally expensive runs. Besides, the WSPSA requires the definition of a weight matrix, which becomes difficult to define as the number of parameters increases. 
In \cite{BayesianOpt, BayesianOptPreprint}, an iterative black box approach is adopted to calibrate an activity-based transport model, with the former applied to a small network (24 zones) and the latter calibrating only 9 behavioral parameters. Paper \cite{MaxLikelihood} succeeds in calibrating 25 behavioral parameters and, similarly to \cite{BerlinMatsim,BerlinMatsim2,Belgium2010}, jointly calibrates an activity-based model with a traffic assignment tool. Papers \cite{BerlinMatsim,BerlinMatsim2} calibrate both the MATSim software and CEMDAP \cite{CEMDAP}, with MATSim receiving the final calibration based on traffic counts. Still, in these works a calibrated traffic assignment model is needed, which increases the overall calibration effort by increasing the number of factors but also by somehow putting the two modules (activity-based and traffic assignment) \enquote{against} each other, in an iterative fashion, whereas a univocal optimization criterion for the two modules is not defined, with convergence being decided by supply-side metrics. This issue is tackled in \cite{MATSimZurich}, where MATSim is integrated with a multinomial discrete choice model considering 12 behavioral parameters. Results show that the choice of behavioral parameters becomes a key element in the simulation pipeline, without which it is not possible to reach a good integration with an ABM while avoiding convergence issues. To address this limitation, \cite{MATSimZurich2} decouples the traffic assignment from the behavioral components; however, the resulting agents' features and the related behavioral constants in a logit model remain not calibrated, which limits the transferability or even the usability of the calibrated model. Overall, MATSim applications, while generally more widespread, are more limited in their ability to predict new technologies and disruptive scenarios while, also, relying on a smaller number of parameters to be calibrated \cite{MATSimBH}.

Based on the above, it is possible to state the following limitations concerning the state-of-the-art:
\begin{enumerate}
\item Existing literature tackling the calibration of behavioral parameters for activity-based models on large-scale scenarios is scarce and only a handful of works try to solve the problem without recurring to heuristics.
\item Most of the calibration methods aim to reproduce outputs of the activity-based models matching the desired supply-side measurements, rather than to calibrate the underlying behavioral parameters. Thus, the calibration of the supply overrules the calibration of the demand.
\item Even the works trying to formalize a rigorous methodology do not consider more than a few dozen of parameters (the maximum number being 98 in \cite{LiThesis}).
\end{enumerate}

\subsection{Bayesian optimization}
BO finds applications in various scientific and industrial domains, e.g., machine learning for hyperparameter optimization \cite{falkner2018bohb, feurer2019hyperparameter}, modeling of population genetics \cite{jarvenpaa2018gaussian}, spreading of pathogens \cite{lintusaari2017fundamentals}, atomic structure of materials \cite{todorovic2019bayesian,zhang2020bayesian}, as well as cosmology \cite{leclercq2018bayesian}, and establishes as a state-of-the-art method in lower-dimensional problems \cite{shahriari2015taking,aushev2020likelihood}. However, the BO performances and its computational efficiency decline as the dimensionality of a problem increases \cite{nott2014approximate,falkner2018bohb,izbicki2019abc,raynal2019abc}, which is the case with the calibration of large-scale ABM that features a large number of behavioral parameters to be tuned.

BO exhibits state-of-the-art performances using Gaussian processes (GP) as a prior distribution to model the surrogate surface and approximate the posterior distribution of the parameters \cite{shahriari2015taking, gutmann2016bayesian}. The probabilistic nature of the GP enables the quantification of the prediction uncertainty based on the spatial vicinity of new samples to the known regions in mathematical spaces. Such quantified uncertainty allows for an efficient trade-off that guides the search for better samples to be sampled, which helps the BO to achieve state-of-the-art performances. However, the GP comes with a computation bottleneck when applied to high-dimensional problems, constituting a setback for their broader adoption in settings of complex parameter spaces \cite{nott2014approximate,falkner2018bohb,izbicki2019abc,raynal2019abc}. Therefore,  the straightforward adoption of the BO method in the calibration of activity-based models is hampered by the large number of parameters to be tuned \cite{ActBMCalib-recent, kuzmanovski2021composite}.

For that purpose, various methods for dimensionality reduction are adopted \cite{blum2013comparative, BayesianOpt, BayesianOptPreprint}, or transformations (including partitioning) of the parameter space are applied \cite{wang2013bayesian, wang2017batched, oh2018bock, mutny2018efficient, nayebi2019framework, kirschner2019adaptive, gramacy2020surrogates}, but neither circumvent the obstacle. Namely, the former requires an increased number of runs, while the latter relies on strong assumptions of low intrinsic dimensionality and compounding effects. \cite{BayesianOpt, BayesianOptPreprint} invest a significant amount of effort to introduce BO in the field of transportation modeling, but fails to do so without an expensive dimensionality reduction using a deep learning methodology, i.e., auto-encoders. The deep learning methodology has been shown in numerous applications to be a valuable approach in high-dimensional spaces, but it is known to be data-intensive \cite{goodfellow2016deep}, which in the context of transportation activity-based models means a greater number of executions of the models.   

Approaches based on transforming the parameter space and decomposing the optimization problem into sub-problems -- each mapped to a lower-dimensional space -- depend on space properties, among which the intrinsic dimensionality is the most important. In that context, previous works explore a latent space where the function is decomposable, either by latent structures \cite{wang2017batched} or additive structures \cite{li2016high, gardner2017discovering}. Another approach to dimensionality reduction involves random projections into a latent space \cite{wang2013bayesian, mutny2018efficient, nayebi2019framework, kirschner2019adaptive} or low-rank matrix approximation \cite{djolonga2013high}. Alternatively, a cylindrical transformation of the parameter space \cite{oh2018bock} and sequential optimization along with a subset of dimensions \cite{li2018high} are adopted in recent studies. However, the authors of these studies report the performance on benchmark optimization functions, showing that the methods perform well on problems with low intrinsic dimensionality, but fail to depart significantly from the initial points in optimization problems with high intrinsic dimensionality.  

Without prior knowledge of the intrinsic dimensionality of the large-scale activity-based models and the corresponding calibration process, we consider a variation of the BO method that uses a dimensionality-wise more robust method to approximate the posterior of the parameters, i.e., Random forests \cite{breiman2001random}. Previously, the method of Random forests has been used in a study with low dimensional optimization problems, featuring discrete mathematical spaces \cite{hutter2011sequential}. 
 
\section{Methodology} \label{Methodology}
\subsection{Activity-based transport models}
An activity-based ABM aims at describing the behavior of potentially millions of agents, each representing a traveler (and/or a vehicle), each one capable of multiple choices through the simulation horizon. 
The decision process for each agent is typically modeled via a nested tree, where each node represents a choice, which is in turn defined via utility maximization, solved via evaluating multiple (utility) functions, each described by several parameters and, crucially, the corresponding weights.
Utility functions are of the form
\begin{equation}
    U_{\textrm{choice}} = f \left( V, \underline{\beta}{}_{\textrm{choice}} \right),
\end{equation}
where $V$ is a set of (known) parameters characterizing the agent, which could include, e.g., the socioeconomic features of each agent (defined while creating the synthetic population), and $\underline{\beta}{}_{\textrm{ choice}}$ are the weights defining how much these parameters concur to the utility function. 
The hundreds of possible combinations among weights result in a problem for which it is impossible to calculate an analytical solution, while the stochastic nature of the utility-maximizing theory makes employing a numerical solution a necessity.
In the following, all weights $\underline{\beta}{}_{\textrm{ choice}}$ present in all utility functions defined in a decision tree are grouped in $\theta$, while the observations are measurements resulting from the emergent behavior of the agents.

\subsection{The conceptual design}
\begin{figure*}[!bp]
    \centering
    \includegraphics[width=0.85\textwidth]{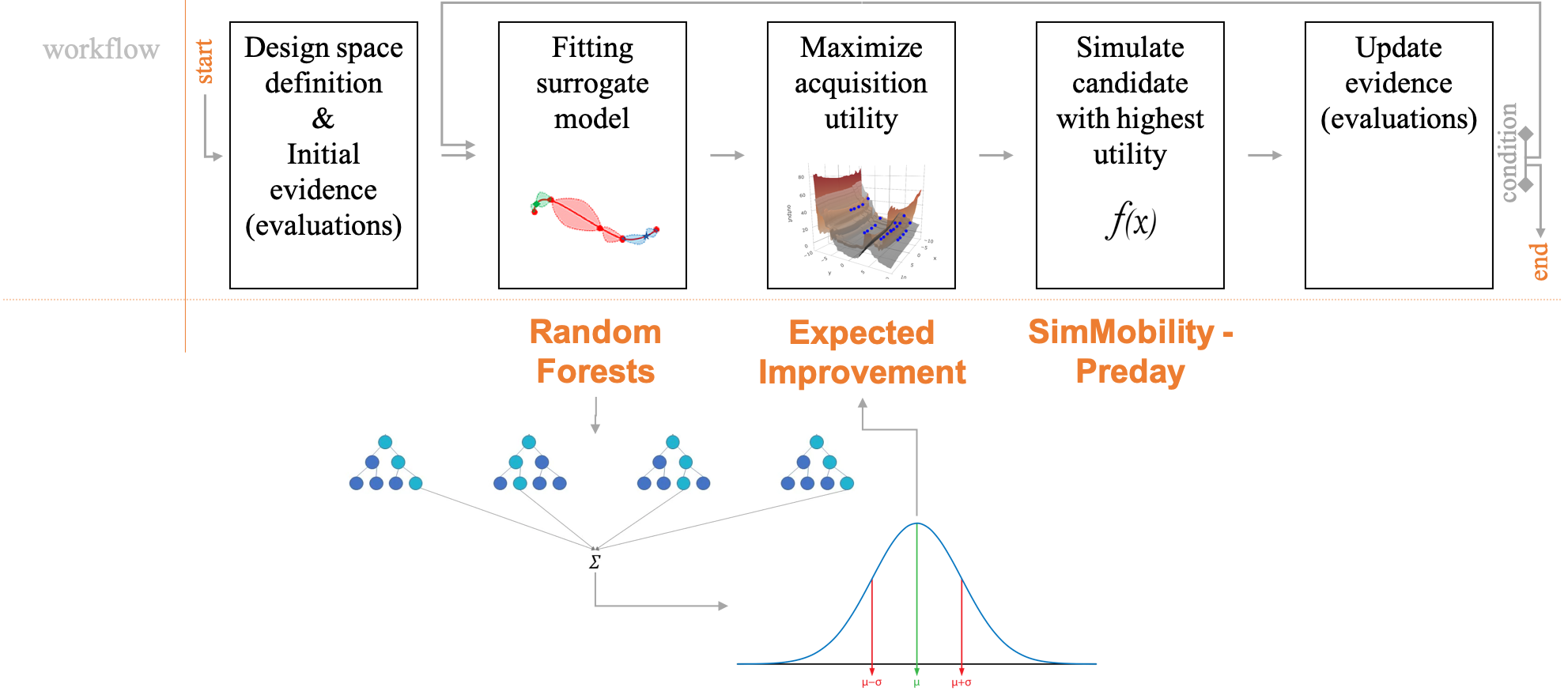}
    \caption{Conceptual design of the iterative Bayesian optimization method with Random Forest as a surrogate model and Expected Improvement (EI) as an acquisition function.}
    \label{fig:soa-bo}
\end{figure*}

BO \cite{shahriari2015taking, gutmann2016bayesian} is a method that seeks the global optima through sequential sampling design and approximation of an underlying likelihood function using a surrogate model over a surrogate (response) surface. The surrogate surface is defined by a parameter space and discrepancy between observed and simulated outputs. The sequential sampling design is an iterative approach, through which new samples (parameters' values) are acquired that maximize the expected (acquisition) utility. The utility is based on an acquisition function, which balances the trade-off between exploration and exploitation of the search space and efficiently guides the search towards the global optima (Fig.~\ref{fig:soa-bo}).

Formally, a simulation model is a generative stochastic process and its calibration corresponds to a statistical inference of a finite number of parameters $\theta \in \mathbb{R}^d$ based on a set of observations $Y_o$:

\begin{equation}
p(\theta|Y_o) = \frac{p(Y_o|\theta) \cdot p(\theta)}{p(Y_o)}, 
\label{equation1}
\end{equation}
where $p(\theta)$ is our prior belief on the distribution of parameter values and $p(Y_o|\theta)$ is the likelihood of the observations, given the parameters, derived from a known function $\mathcal{L}(\theta)$. Since the analytical form of $\mathcal{L}(\theta)$ is unknown in the underlying challenge, we use the notation $L(\theta)$ that needs to be approximated over a set of $N$ samples - $\tilde{L}^N(\theta)$. The notation is simplified if the marginal distribution $p(Y_o)$ is omitted because it does not depend on $\theta$:

\begin{equation}
p(\theta|Y_o) \propto L(\theta) \cdot p(\theta),
\end{equation}

where the $L(\theta)$ is approximated over a finite sample set ($\tilde{L}^N(\theta)$) and reconstructed as the number of samples increases:

\begin{equation}
\lim_{ N \to \infty } \tilde{L}^N(\theta) = L(\theta).
\end{equation}

\subsection{The surrogate function and the sampling design}
The approximation ($\tilde{L}^N(\theta)$) of the likelihood function ($L(\theta)$) is the formal task we address with the BO methodology, using a surrogate function and the sequential sampling design \cite{gutmann2016bayesian}. To model the surrogate function, we use a regressor such as Random forests (RF) \cite{breiman2001random}, which is used to estimate the acquisition utility of newly sampled parameter values through the \textit{Expected Improvement} (EI) as an acquisition function \cite{movckus1975bayesian}:

\begin{equation}
    EI(\theta|\mu,\sigma,f^\textrm{*}) = \sigma(\theta)[z\Phi(z) + \phi(z)]
    \label{eq:ei}
\end{equation}
\begin{equation}
    z = \frac{f^\textrm{*} - \mu(\theta)}{\sigma(\theta)},
\end{equation}

\noindent where $\sigma(\theta)$ and $\mu(\theta)$ are statistics of the inferred posterior distribution, $f^*$ is the active optima discovered in the previous iterations, and $\Phi$ and $\phi$ are probability density and cumulative distribution function in terms of the standard normal distribution, respectively. The expected improvement $EI(\theta) = 0$ if $\sigma(\theta) = 0$. The analogy behind (\ref{eq:ei}) reveals the exploration-exploitation trade-off that favors larger uncertainties that are close to the latest discovered optimal region(s).

The RF \cite{breiman2001random} is an ensemble method composed of $C$ regression trees. Regression trees follow the decision tree concept, with a structure of decision binary nodes built iteratively in a top-down fashion. Each regression tree is built over a subspace of the parameter space, designed by random subsets of both the features (dimensions) and bootstrap samples. Therefore, given a dataset, each regression tree predicts the target for a specific region in the defined space. The prediction of the ensemble, on the other hand, is an aggregation (average) of the outcomes of all $C$ tree base predictors:

\begin{equation}
    \mathcal{RF}(\theta | \Theta,Y) = \frac{1}{C}\sum_{i=1}^{C} \tau_i;~~ \tau_i = T_i\left(\theta | \Theta_{i},Y_{i}\right),
\label{eq:rf}
\end{equation}

\noindent where $C$ is the number of tree predictors, $\Theta_{i}$ and $Y_{i}$ are training datasets of $i$-th regression tree $T_i$ that provides a prediction $\tau_i$, while $\Theta$ and $Y$ are global training dataset and the corresponding label set, respectively.

The RF method has a small number of hyper-parameters that can significantly influence the outcome, among which most commonly tuned are the number of tree components $C$, the minimum number of samples in a terminating node that controls the structure growth and over-fitting settings of the individual tree components, and the number of features to design a sub-space or partition. Additionally, the RF method has shown excellent robustness over high-dimensional data, which limits the bias of the overall predictions by maximizing the variance between base predictors \cite{friedman2007bagging}. 

However, in the context of BO, RF models lack: (i) uncertainties in the quantification of predictions (non-probabilistic output); and (ii) predicting a value outside of the observed range. Thus, as a standalone surrogate model, the RF greatly affects the efficiency of a probabilistic acquisition function (e.g., EI) in acquiring new promising samples \cite{hutter2011sequential,shahriari2015taking}.
\subsection{The improved Random Forest}
In order to comply with the expected probabilistic output, the RF method is adapted so that it approximates a parametric (normal) probability distribution of the evaluated input parameter values ($\theta$), through empirically derived mean ($\mu_{\theta}$) and standard deviation ($\sigma_{\theta}$):

\begin{gather}
    \mu_{\theta} = \mathcal{RF}(\theta | \Theta,Y) = \frac{1}{C}\sum_{i=1}^{C} \tau_i; \\ 
    \sigma_{\theta} = \sqrt{\frac{1}{C}\sum_{i=1}^{C} \left(\mu_{\theta} - \tau_i\right)^2},
\end{gather}

\noindent where $\tau_i$ corresponds to a prediction of a single decision tree model in the RF, as described in Eq.~\ref{eq:rf}.

As we adapt the RF into a compatible method for modeling the surrogate function that enables estimation of the acquisition utility for each new sampled parameter value, a component of BO, yet to be formalized in our calibration framework, is the optimization of the acquisition utility at each iteration of the iterative optimization process. The optimization of the acquisition utility corresponds to finding a set of parameters values ($\theta^{\textrm{*}}$) that maximizes the utility:
\begin{equation}
    \theta^{\textrm{*}} = \underset{\theta}{\textrm{argmax}}\, EI\left(\mu_{\theta}, \sigma_{\theta},\theta_{\textrm{prev}}^{\textrm{*}}\right),
\end{equation}
where $\theta_{prev}^{\textrm{*}}$ is the optimal set of parameter values obtained in previous iterations of the process.

Finding the set of parameters' values that maximizes the acquisition utility (estimated to perform best) can be performed in various ways, including Random search, Thompson sampling, gradient-based, or population-based (evolutionary) optimization methods \cite{shahriari2015taking, gutmann2016bayesian, feurer2019hyperparameter}. In this study, we examined the performances of the Random search, gradient-based, and population-based methods, observing that the gradient-based outperforms the rest. Therefore, for our case study, we adopt the gradient-based Limited-memory Broyden–Fletcher–Goldfarb–Shanno algorithm with box constraints (L-BFGS-B) \cite{byrd1995limited,nocedal1999numerical}.  

\subsection{The selection of the best-performing simulation}

The stochastic nature of the proposed method requires that the optimization is performed multiple times, by which the possibility to found a locally optimal solution is eliminated. However, depending on the complexity of the high-dimensional problem at glance and definition of the optimisation objectives (cost), it may be not feasible to completely avoid locally optimal regions of the problem space, as there may be more of them. Therefore, \textit{ex post} Pareto analyses of simulated candidates according to a broader set of criteria is recommended. These shall represent some important benchmarks or margins of the considered case study and have to be defined accordingly. They differ from the optimisation objectives (performance measure) that is exploited to guide the algorithm through the learning process, and they are used to filter out  non-robust simulations in accordance to the Pareto objectives.

\section{Case Study}

\begin{figure}[h]
\centering
\includegraphics[scale=0.11]{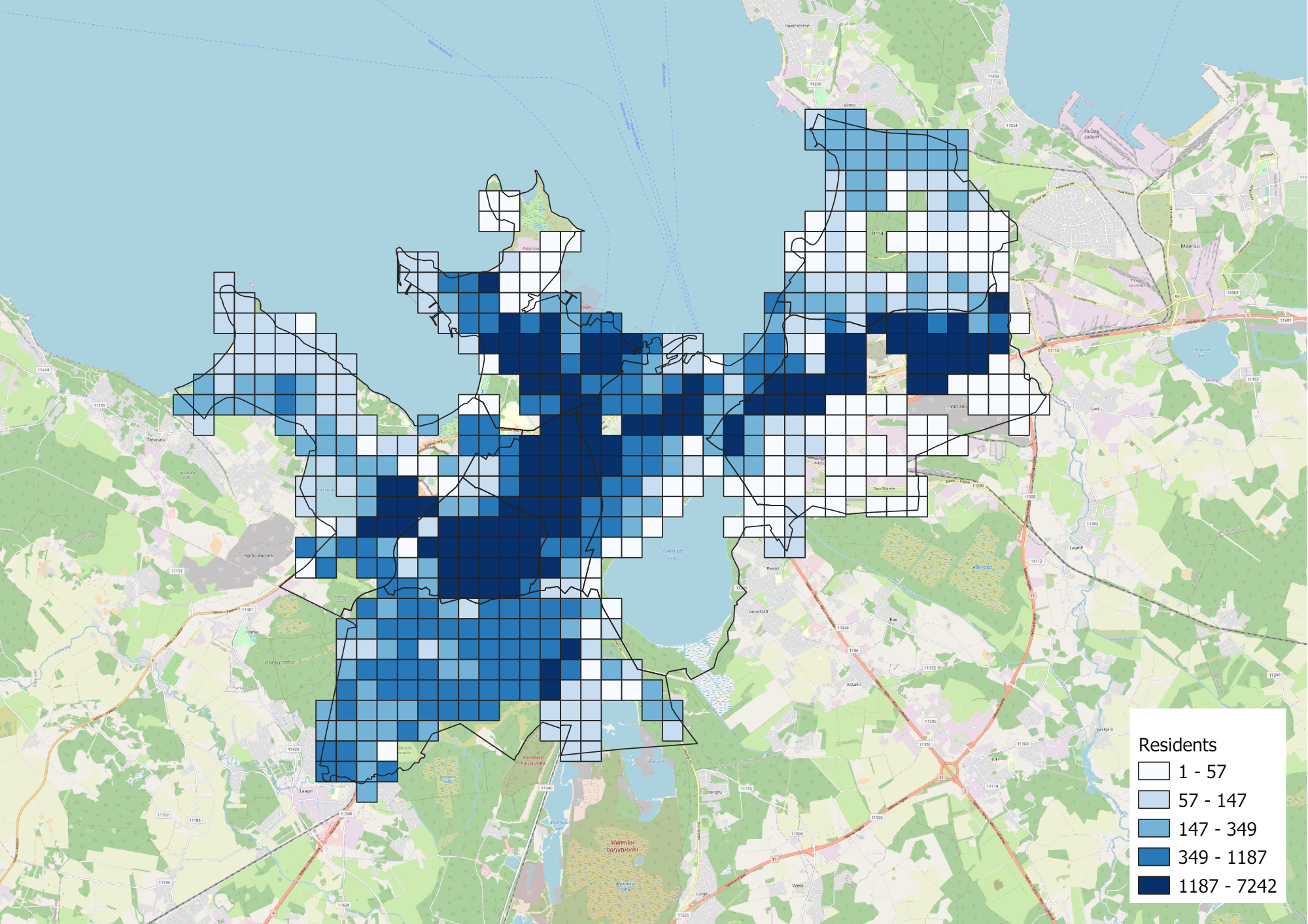}
\caption{Spatial distributions of residents}
\label{fig:tallinn}
\end{figure}

The proposed methodology is tested for modeling the city of Tallinn (Estonia). The reference year is 2015 as the mobility survey used both for the generation of the database population and for the calibration was carried out in the said year (Fig.~\ref{fig:tallinn}). SimMobility MT - Preday \cite{SimMobMT} is employed as the activity-based model, which takes as input a Postgres database containing the following features:
\begin{enumerate}
\item A population of $\sim$400.000 synthetic individuals, matching the city's whole population, while each individual is characterized by socioeconomic features such as age, gender, income, workplace, etc.
\item A spatial map of 616 zones, each 500x500 meters wide (Fig.~\ref{fig:tallinn}).
\item Skim tables detailing the costs of different travel modes, including waiting time, time on board, etc., among the different zones.
\end{enumerate}

Four transport modes are considered: Public Transport, Private Vehicles, Walking, and Others (e.g., motorcycles). The spatial resolution has been set to 500 m to realistically capture the Walking mode of transport, while the skim matrix for public transport has been extracted through Open Trip Planner \cite{OTP}. More details about the synthetic population, its generation, and assignment, may be found in \cite{SynthPop}. 

A run of Preday results in a set of Daily Activity Schedules, one for each leg of a tour. An extract is provided in Table~\ref{tbl:DAS-107} for a randomly chosen individual (ID 107).

\begin{table*}[h!]
\begin{center}
\caption{The daily activity schedule for individual 107}
\resizebox{\textwidth}{!}{
\begin{tabular}{||c c c c c c c c c c c c||} 
 \hline
 \textbf{person id} & \textbf{tour no} & \textbf{tourType} & \textbf{stop no} & \textbf{stop type} & \textbf{stop location} & \textbf{stop mode} & \textbf{primary stop} & \textbf{arrival time} & \textbf{departure time} & \textbf{prev stop location} & \textbf{prev stop departure time}\\
 \hline\hline
 107-1 & 1 & Education & 1 & Education & 166 & BusTravel & True & 7.25 & 13.75 & 569 & 7.25\\ 
 \hline
 107-1 & 1 & Education & 2 & Home & 569 & BusTravel & False & 13.75 & 15.75 & 166 & 13.75\\ 
 \hline
 107-1 & 2 & Work & 1 & Work & 415 & BusTravel & True & 15.75 & 18.25 & 569 & 15.75\\
 \hline
 107-1 & 2 & Work & 2 & Home & 569 & BusTravel & False & 18.25 & 19.25 & 415 & 18.25 \\
 \hline
 107-1 & 3 & Shop & 1 & Shop & 496 & Walk & True & 19.75 & 23.25 & 569 & 19.25\\
 \hline
 107-1 & 3 & Shop & 2 & Home & 569 & Walk & False & 23.75 & 26.75 & 496 & 23.25\\
 \hline\hline
\end{tabular}}

\label{tbl:DAS-107}
\end{center}
\end{table*}
The main strength of Preday lies in the behavioral models used, i.e., a series of nested logit functions allowing to simulate the travel demand based on an established methodology \cite{SimMobMT,Logsum,BostonSimMob}. This feature of the model allows simulating future scenarios for which the ground data (e.g., traffic counts) are not yet available, which is done via computing utilities at different levels (binary choice to leave the residence, type and number of tours, modes and destinations, time of the day, and stops). These choices are interrelated, an aspect that is accounted for through the computation of logsums (defined in \cite{Logsum} as the log of the denominator of a logit choice probability). 

In the following, the formulation of utility for the binary choice to leave or not the residence (top of the nested logit tree) and the mode choice (bottom of the nested logit tree) are reported (\ref{eq:Ubinary}),~(\ref{eq:utilitymode}) to highlight the high number of behavioral parameters involved (and, crucially, to be calibrated).
\begin{align}
U_\textrm{binary} = f(&V_{\textrm{case\_study}},\underline{\beta}_{\textrm{~female\_travel}}, \underline{\beta}_{\textrm{~age\_category}},\nonumber \\
&\underline{\beta}_{\textrm{~children\_in\_household}},\underline{\beta}_{\textrm{~income}},\underline{\beta}_{\textrm{~missing\_income}}, \nonumber \\
&\underline{\beta}_{\textrm{~work\_at\_home}},\underline{\beta}_{\textrm{~number\_of\_cars\_in\_household}}, \nonumber \\
&\underline{\beta}_{\textrm{~dptour\_logsum}},\underline{\beta}_{\textrm{~employment\_status}})\label{eq:Ubinary}
\end{align}
Note that each variable in (\ref{eq:Ubinary}) is a vector of \textbeta s, including as many behavioral variables as categories considered. For example, \underline{\textbeta} for age category is a vector with 5  \textbeta s, since 5 are the age categories considered. Overall, binary alone includes 25 \textbeta s to be calibrated. For further information about the nested logit tree, please refer to \cite{SimMobMT}. 

\indent On the other side of the tree, the utility related to the bus mode is computed as:
\begin{align}
U_\textrm{bus} \!=\! f(&V_{\textrm{case\_study}},\underline{\beta}_{\textrm{~cons\_bus}}, \underline{\beta}_{~\textrm{tt}}, \underline{\beta}_{\textrm{~walk\_time}}, \underline{\beta}_{\textrm{~wait\_time}},\underline{\beta}_{~\textrm{cost}},\nonumber\\
&\underline{\beta}_{\textrm{~cost\_over\_income}}, \underline{\beta}_{\textrm{~central\_district}},\underline{\beta}_{~\textrm{transfer}},\underline{\beta}_{\textrm{~female\_num\_of\_cars}},\nonumber \\
&\underline{\beta}_{\textrm{~number\_of\_cars\_in\_hh}},\underline{\beta}_{\textrm{~agecat\_num\_of\_cars}})
\label{eq:utilitymode}
\end{align}
Also in this case, the above is a simplified version for presentation purposes and the number of \textbeta s required to compute U\textsubscript{bus} is 18. When four modes and four levels of the nested logit tree are considered, the resulting total number of  behavioral parameters (\textbeta s) to be calibrated is 477. The whole set of \textbeta s represents the $\theta$ parameters described in Section~\ref{Methodology}, they constitute the search space explored by the algorithm. The full list of \textbeta s is publicly available\footnote{\url{https://github.com/Angelo3452/Tallinn-Synthetic-Population/tree/main/SimMobility\%20MT\%20Database/\\Postgres\%20database}}. 

The calibration is carried out against the following baseline data:
\begin{enumerate}
\item OD matrix at subdistrict level. Tallinn has 82 subdistricts and the movements across them at each time of the day are extracted and upscaled from a mobility survey obtained from Taltech University \cite{MobilitySurvey}.
\item Statistical margins concerning workplace distributions and totals at the cell level (500x500m). The methodology behind these is detailed in \cite{SynthPop}. A similar method, albeit simplified, was applied for the margins concerning school institutions. 
\item Modal share, as detailed in \cite{ModalShare}.
\end{enumerate}

\noindent These represent the observations~$Y_o$ introduced in (\ref{equation1}) and the achieved match are detailed in the calibration and results section.

\section{Calibration and results}

\begin{table*}[h!]
\begin{center}
\caption{Hyperparameters used for the algorithms utilized in this study. For the hyperparameters not included in this table, default values are used.}
\begin{tabular}{||c c c||} 
 \hline
 \textbf{Algorithm} & \textbf{Hyperparameter} & \textbf{Value(s)} \\
 \hline\hline
 Bayesian optimization & Termination conditions & 500 iterations or 500 hours \\ 
 \hline
 Bayesian optimization & Initial sampling & Latin Hypercube Sampling \\ 
 \hline
 Bayesian optimization & Retrain surrogate model & every 5 iterations \\
 \hline
 Bayesian optimization & Acquisition optimization & L-BFGS-B \cite{byrd1995limited,nocedal1999numerical} \\
 \hline
 Random forests & Number of trees & 1000, 3000, 5000 \\
 \hline
 L-BFGS-B & Termination conditions & 1000 iterations \\
 \hline 
 L-BFGS-B & Step sizes for approximation to gradient & 0.5 \\
 \hline\hline
\end{tabular}
\label{tbl:hyperparam}
\end{center}
\end{table*}

\begin{figure}[tb]
\centering
\includegraphics[scale=0.43]{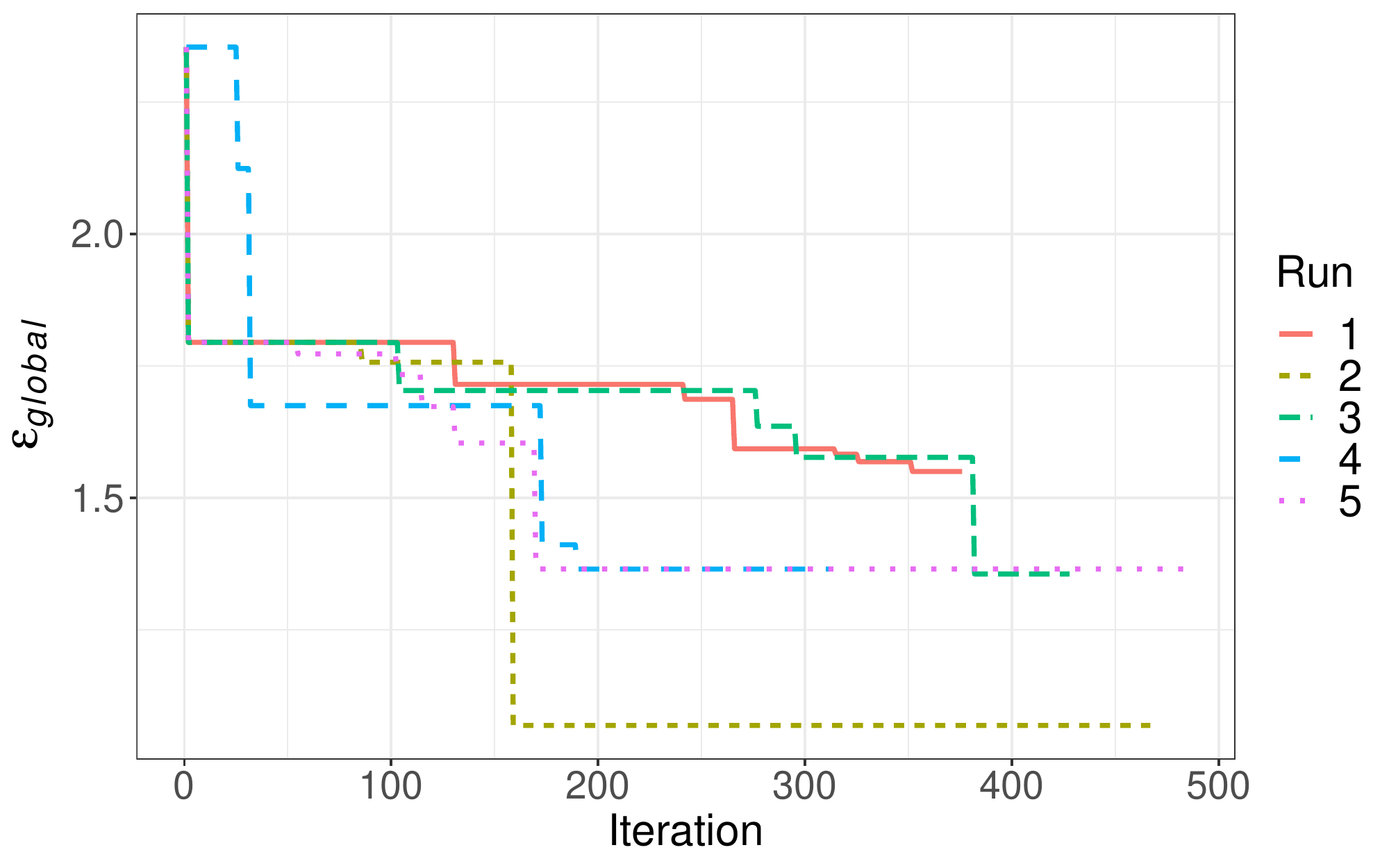}
\caption{Performance measure progression across 500 iterations for 5 different runs }
\label{fig:Performance measure}
\end{figure}

\begin{figure}[tb]
\centering
\includegraphics[scale=0.43]{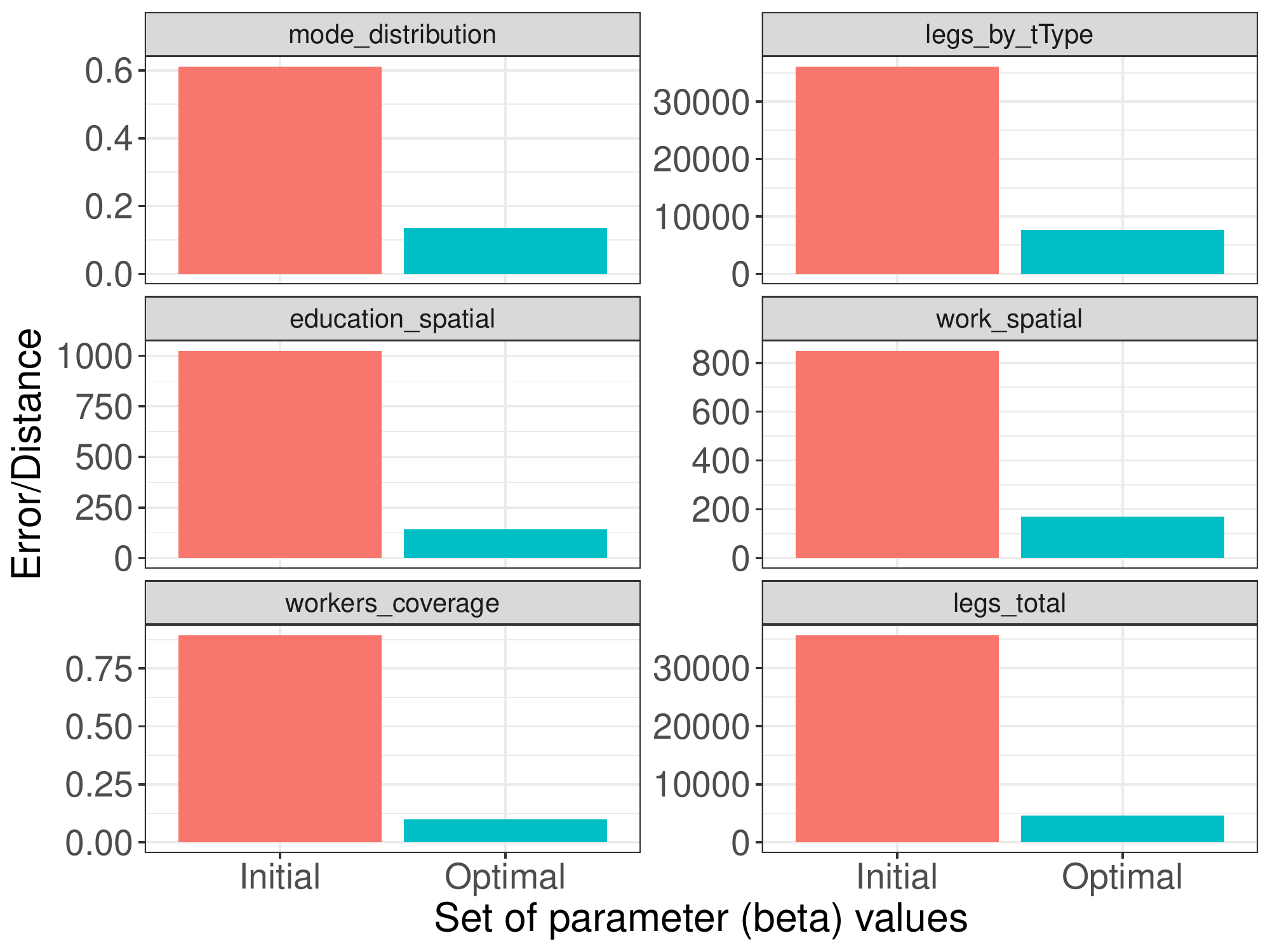}
\caption{Comparison between the starting point (1) and the best simulation in run~2~(182) -- benchmarks against the baseline and residual error; the y-axis represents the percentage for worker\_coverage and mode distribution, and the number of legs for all the other benchmarks}
\label{fig:Best simulations}
\end{figure}

\begin{figure}[tb]
\centering
\includegraphics[scale=0.45]{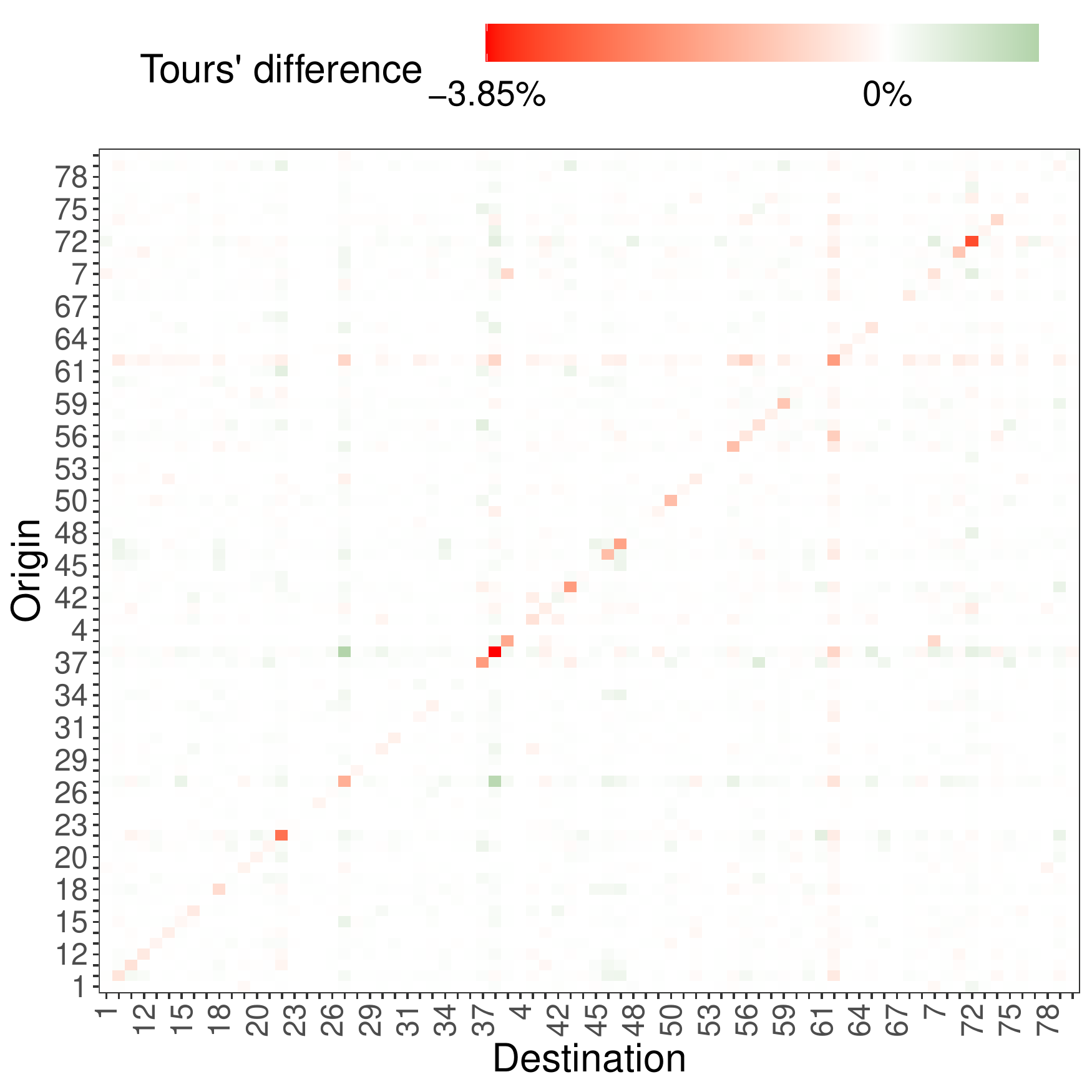}
\caption{Difference between the number of trips for each OD pair, calculated between the simulated ones and the total obtained instead by upscaling the mobility survey to the whole population. The axis labels include every third district.}
\label{fig:OD difference}
\end{figure}

\begin{table*}[h!]
\begin{center}
\caption{Numerical comparison between the best-performing simulation 182 in run 2 and the baseline}
\begin{tabular}{||c c c c c c||} 
 \hline
 Variable & Category & Simulated & Baseline & Difference & Percentage Error\\ [0.5ex] 
 \hline\hline
 legs [abs] & Education & 19259 & 19563 & -304 & 1.55 \\ 
 \hline
 legs [abs] & Total & 117093 & 112481 & 4612 & 4.10 \\
 \hline
 legs [abs] & Work & 44521 & 36788 & 7733 & 21.02 \\
 \hline
 mode [\%] & car & 0.420 & 0.488 & -0.069 & - \\
 \hline
 mode [\%] & other & 0.014 & 0.012 & 0.003 & -\\
 \hline
 mode [\%] & PT & 0.366 & 0.256 & 0.110 & -\\
 \hline
 mode [\%] & walk & 0.2 & 0.239 & -0.039 & -\\ [1ex] 
 \hline
\end{tabular}

\label{table:1}
\end{center}
\end{table*}

\begin{figure}[h]
\centering
\centerline{\includegraphics[scale=0.31, left]{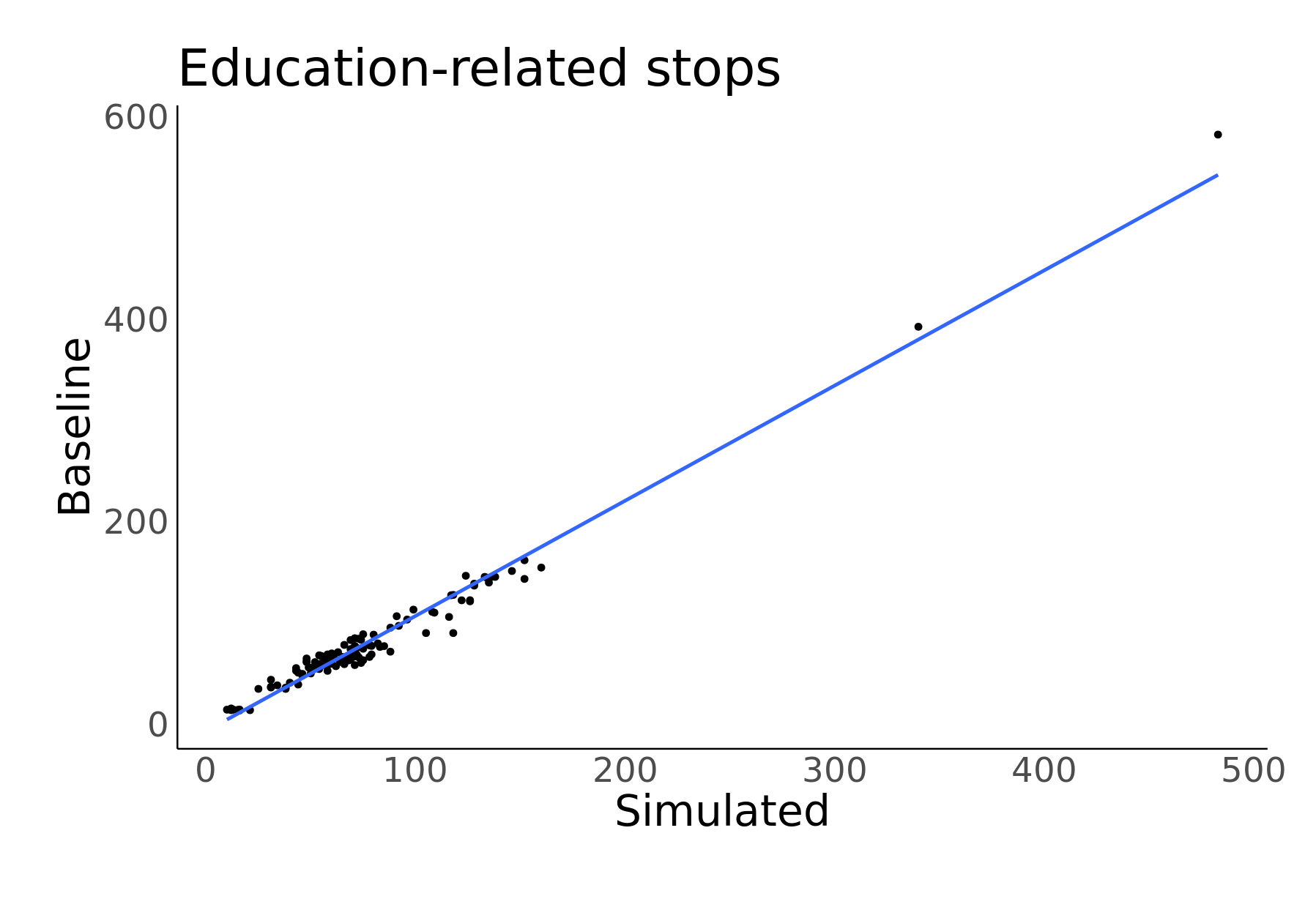}}
\centerline{\includegraphics[scale=0.31, left]{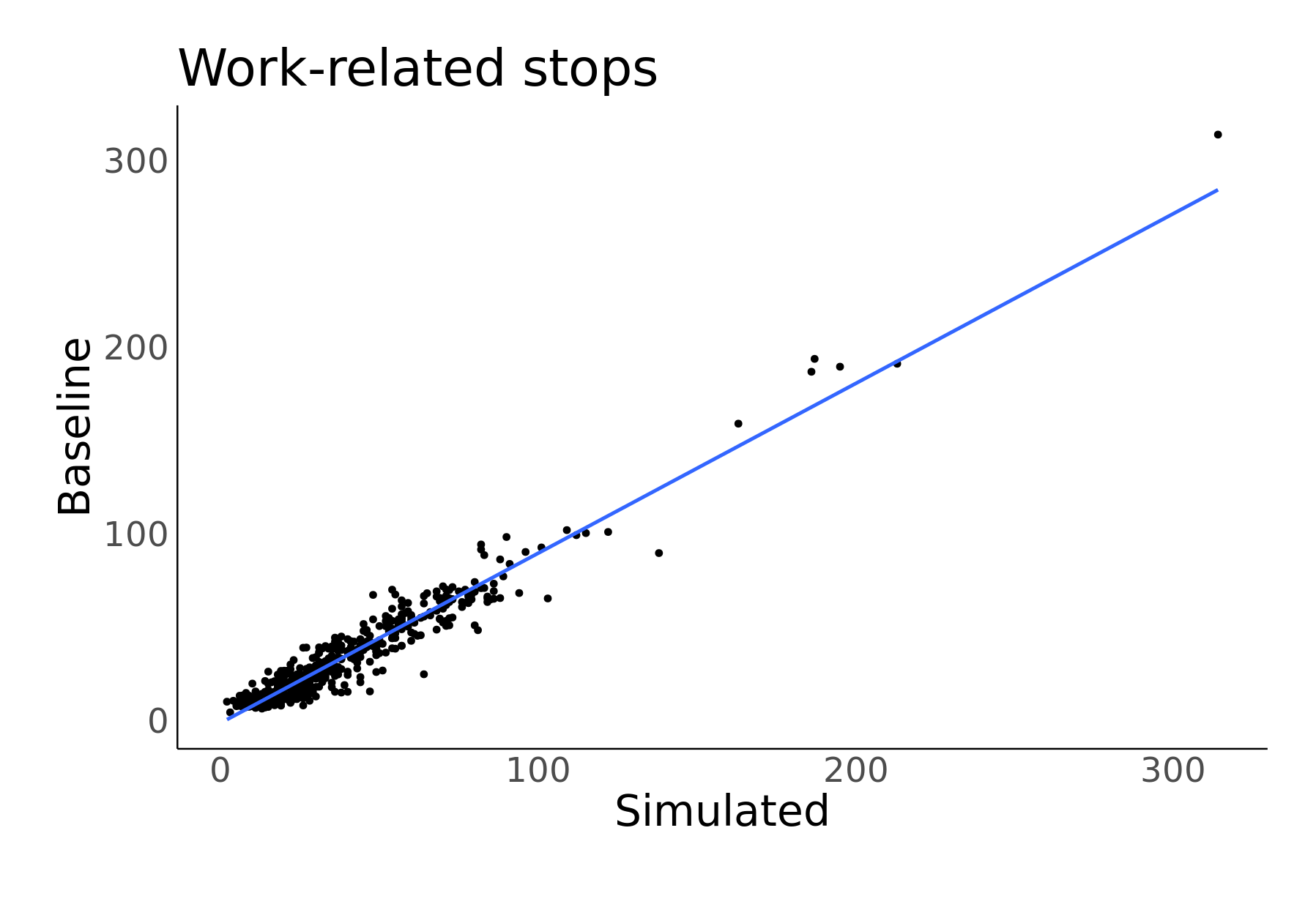}}
\caption{Comparison of attendances at anchor points for education-related and work-related reasons; the x- axis represents the number of trips to the anchor point in the calibrated simulation, the y- axis represents the baseline number of trips to the anchor point from the baseline}
\label{fig:Anchor points}
\vspace{-2mm}
\end{figure}

\begin{figure}[tb]
\centering
\includegraphics[scale=0.15]{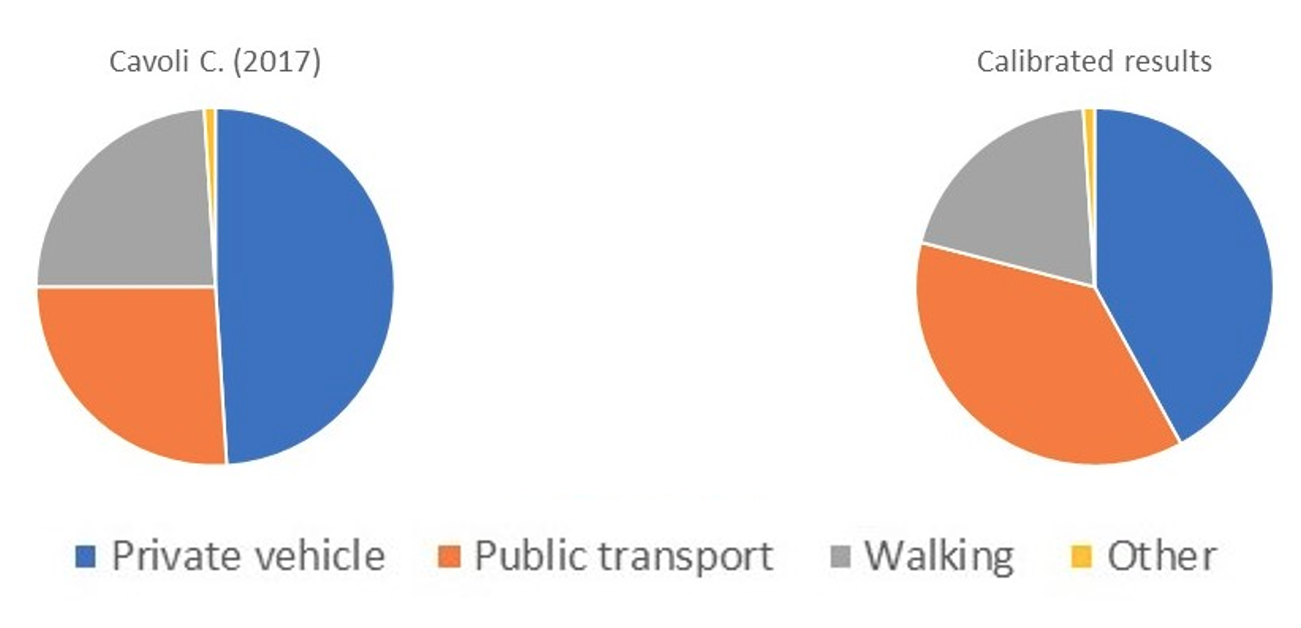}
\caption{Modal share comparison with \cite{ModalShare} }
\label{fig:Modal share}
\vspace{-4mm}
\end{figure}

\begin{figure}[tb]
\centering
\includegraphics[scale=0.33]{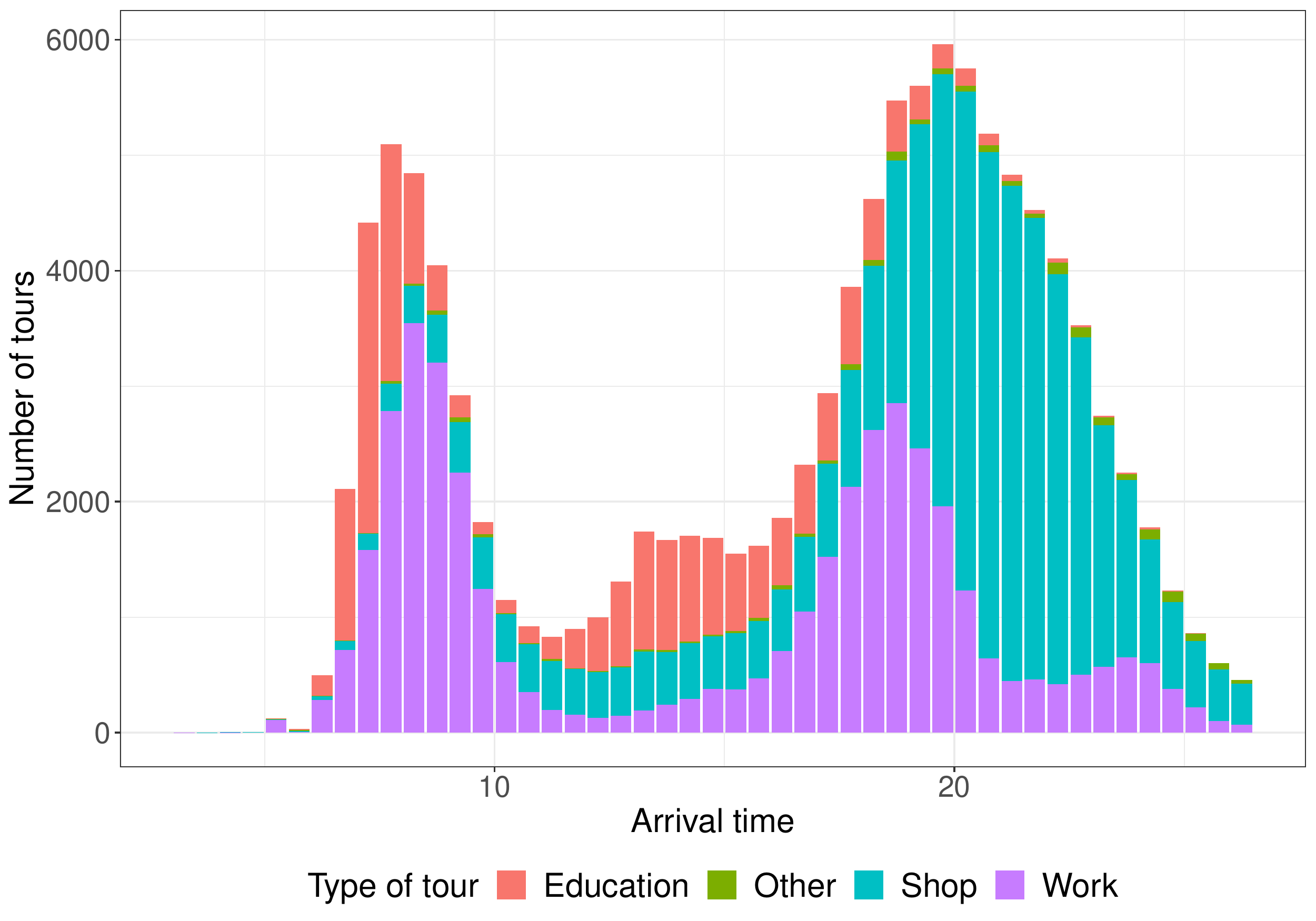}
\caption{Temporal distribution of tours throughout the day for the best-performing simulation 182 in run 2}
\label{fig:Tours through the day}
\end{figure}

\begin{figure*}[tb]
\centering
\includegraphics[scale=0.51,clip]{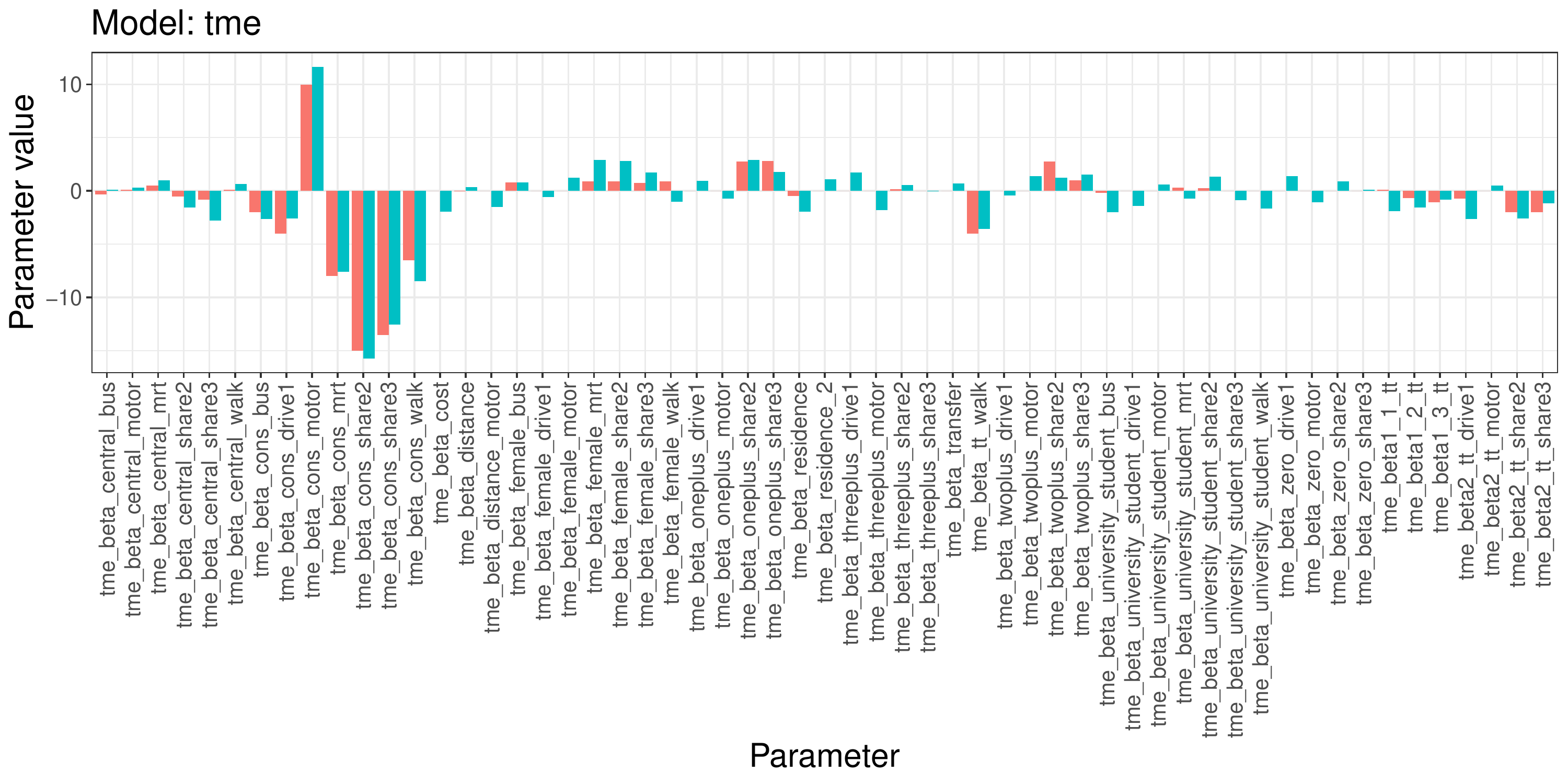}
\caption{Comparison of the parameters -- starting (red) and final (green) values -- in the logit branch modeling the mode choice for trips to school or other educational institutions}
\label{fig:Betas}
\end{figure*}

\subsection{The objective function}
The design of the calibration process with the proposed methodology features a)~a custom objective function, b)~an iteration process for the calibration runs, and c)~the definition of hyperparameters for both optimization methods, i.e., the global objective and the inner acquisition function. 

Formally, the calibration of a simulation model is an optimization task that aims to minimize the discrepancy between a set of simulated and observed outputs. The discrepancy is measured via an objective function and, within this study, we compile a custom function that comprises three components: (i) Origin-Destination (OD) matrix; (ii) share of transportation modes; and (iii) coverage of the employed individuals with scheduled work tours.   
All three components are adjusted to result in similar ranges and as such share equal contributions to the final objective function:
\begin{equation}
    \epsilon_{\textrm{global}} = \epsilon_{\textrm{od}} \cdot \epsilon_{\textrm{$\mathcal{M}$}} \cdot \epsilon_{\textrm{workers}} 
\end{equation}
\begin{equation}
    \epsilon_{\textrm{od}} = \frac{1}{100}\sqrt{\frac{1}{n}~\sum_{i=1}^{n}\sum_{j=1}^{n} \left(od_{ij} - \tilde{od}_{ij} \right)^2 }
\end{equation}
\begin{align}
    \epsilon_{\textrm{$\mathcal{M}$}} = 1 + \sqrt{\sum_{i \in \mathcal{M}} \left(\mathcal{M}_{i} - \tilde{\mathcal{M}}_{i} \right)^2 }
\end{align}
\begin{align}
    \mathcal{M} = \{\text{public}, \text{car}, \text{walk}, \text{other}\}
\end{align}
\begin{equation}
    \epsilon_{\textrm{workers}} = 2 - w_{\textrm{assign}}/w_{\textrm{total}},
\end{equation}
\noindent where $n$ is the number of districts covered within the OD matrix, $od_{ij}$ and $\tilde{od}_{ij}$ are numbers of observed and simulated tour legs for a given element of the OD matrix, respectively, where~$i$ is the origin and~$j$ is the destination. $\mathcal{M}_i$ and $\tilde{\mathcal{M}}_i$ are shares of observed and simulated transportation modes, respectively, whereby, for both, holds $\sum_{i \in \mathcal{M}} \mathcal{M}_i = 1$ and $\sum_{i \in \mathcal{M}} \tilde{\mathcal{M}}_i = 1$. Regarding the workers component, $w_{\textrm{assign}}$ corresponds to the number of employed individuals with scheduled at least one work-based tour, and $w_{\textrm{total}}$ is the total number of employed individuals in the population. 

\subsection{The iterative algorithm}
In order to account for the stochasticity of the Bayesian optimization and produce stable results, the calibration process is repeated multiple times (five in our experiments), with different random seeds and the initial dataset. The starting values of the behavioral parameters have been set within a realistic range but this preadjustment does not amount to a pre-calibration, as the results from the first simulation reported in Figs.~\ref{fig:Performance measure} and~\ref{fig:Best simulations} show. The reported results are summarized across all runs, with the selected optimal solution (run no. 2) outperforming all other runs.

Each independent run is performed with the same hyperparameters for the optimization methods. A summary of all hyperparameters used in this study is presented in Table~\ref{tbl:hyperparam}.

\noindent In Fig.~\ref{fig:Performance measure}, the progression of the performance measure for 5 sets of iterations is reported.

In order to investigate broader aspects of the solutions and confirm their robustness, we examine the Pareto front of all potential solutions through six additional criteria: i)~the modal share ($\mathcal{M}_{i} - \tilde{\mathcal{M}}_i< 10\%$), the spatial distribution of ii)~work and iii)~education trips, iv)~the absolute number of workers, the v)~total legs, and vi)~the spatially distributed legs (note that the last two items are directly derived from the OD matrix). Hence, the defined cost function guides the algorithm in its optimization process, but the final results are analyzed and a selection is made against a broader set of criteria.
The post analysis confirms that run number 2 outperforms the others after slightly more than 150 iterations and reaches a performance value $ \epsilon_{\textrm{global}} \approx 1.06$. 
\subsection{Calibration against the baseline}
The results reported in the following are the ones arising from run 2. The overall improvement is evident in Figs.~\ref{fig:Performance measure} and~\ref{fig:Best simulations}. Fig.~\ref{fig:Best simulations} shows the best-performing simulation in run 2, namely 182. It is compared with the initial simulation which results instead in 0 satisfactory benchmarks and high errors across all the measurements. The second and third benchmarks (legs\_by\_rType and educational spatial, not perfectly met in 182) reflect instead absolute quantities and, while the latter is quite small and can be considered a match, the former will be further commented on in the following. We would like to stress that these six criteria differ from the metrics selected for the performance measure. In fact, while the latter guides the algorithm and the learning process, the former ones are exploited to filter the best simulations after each iteration.
 
While Fig.~\ref{fig:Best simulations} and Table~\ref{table:1} report only the error in the total number of legs, their distribution across the 82 subdistricts of Tallinn has also been compared against the baseline (i.e., the mobility survey). The algorithm is remarkably able to reproduce correct spatial distributions as in Fig.~\ref{fig:OD difference}.

It should be stressed that these matrices cover a period of 24 hours. This means that, as in Fig.~\ref{fig:OD difference}, the absolute error of private vehicle trips outside the diagonal averages at 15.92 vehicles. If one considers that the magnitude of simulated trips by private vehicles settles around 50.000 trips (each simulation runs over 10\% of the population), the match between the two matrices is impressively close. 
The diagonal does instead show a somewhat higher variance (albeit still within a reasonable margin, an absolute average of 267.82 trips) and all the cells appear to be underestimated in the number of intrazonal trips. Still, intrazonal trips are commonly more difficult to frame, so a higher error was expected.
The algorithm succeeds also in framing the tour types and spatial distribution of work and education trips, somehow trickier because mandatory, thus subject to stricter correspondence against the benchmark. Table~\ref{table:1} reports a comparison of the totals while Fig.~\ref{fig:Anchor points} reports the spatial distributions of school and work trips. 

As can be seen, the algorithm faithfully frames both education and totals, while slightly overestimating work tours. The overestimation has been considered less of a problem than an underestimation since it was considered mandatory that all the work trips would be framed (every employed person goes to the workplace). An overestimation of work trips does instead mistake only the aim of a tour, this is probably caused by the \enquote{emphasis} that the performance measure puts on simulating all the employees going to their workplace. 

The distribution of work and education destinations modeled has been checked against the workplaces and education anchor points previously assigned to each eligible individual in the synthetic population \cite{SynthPop}. Fig.~\ref{fig:Anchor points} shows a good match between the two, which implies that the calibrated utility parameters do indeed result in the right number of trips and spatial distribution for these two categories. Modal share also reaches a reasonable level of precision against the share recorded in \cite{ModalShare}, as shown in Fig.~\ref{fig:Modal share}. 

Another important result of the calibration is the time distribution of the stops in each tour throughout the day. This is reported in Fig.~\ref{fig:Tours through the day}. 
As mentioned, the case study has 2015 as the reference year (ante COVID-19 pandemic). This means that the usual travel pattern showing two peaks (one in the morning and one in the afternoon) was to be expected and is coherent with the case study. Besides, the model clearly captures the different dynamics such as an education peak in the early afternoon or a spike in leisure trips in the evening (after work).

While most of the tours do fall in a realistic pattern, the calibrated behavioral parameters result in a small percentage of tours (4\% percent) allocated in the last available time slot. This is bound to how SimMobility Preday models the time of the day, allocating at the very end all the tours that could not be fitted through the day.

\subsection{Analysis of the calibrated parameters} 
The set of results provided in this section and their comparison with baseline values should clarify how the algorithm reaches an acceptable solution in a completely automated way. It is important to stress how this calibration process differs from the traditional one for Preday, which is carried out manually by tuning the various parameters\footnote{\url{https://github.com/smart-fm/simmobility-prod/wiki/Mid-term-calibration}}, improving its results.
Besides, in the literature review reported above, it was highlighted how other, more complex methods, do not encompass as many \textbeta s as the proposed algorithm. Fig.~\ref{fig:Betas} provides a snapshot of the behavioral parameters and their original and final values for one level of the nested logit tree. For the complete list of \textbeta s, please refer to the publicly available repository\textsuperscript{1}. 
One fundamental aspect should be highlighted: the algorithm allows to expand the set of parameters that are calibrated, as it appears in the tme branch of the logit tree (the branch addressing modal choice for education) plot in Fig.~\ref{fig:Betas}, where many parameters have non-null values only for the calibrated results. This is because the starting values (manually defined) could not encompass such a large set of behavioral parameters that were then set to 0.

\section{Discussion and conclusions}
The paper presented a new algorithm to calibrate a large number of parameters by exploiting a surrogate model and BO techniques, applied to a case study to prove the effectiveness of the proposed method in calibrating hundreds of behavioral parameters for an activity-based model. The result shows a satisfactory match between the modeled outputs and the baseline, built from an available mobility survey and aggregate data. By calibrating the model through the presented algorithm, it was possible to tune a wider set of behavioral parameters than it would have been manually or through heuristics. As shown in the literature review, no other work succeeded in calibrating as many as 477 parameters, although avoiding doing so would strongly reduce the effectiveness of a nested logit model detailed enough to consider different socio-demographic features. The algorithm searches for the best-performing solution by perturbing all 477 behavioral parameters (\textbeta s through the paper). This is a task that could hardly be performed by hand. 

By automating the process and exploiting a surrogate model, the algorithm bypass the need to set up and run multiple runs of the activity-based model. To frame the computational effort that would be required, a single run of SimMobility Preday for a population of $\sim$400.000 individuals takes around 6-8 hours on a laptop (Intel core i5 - 1.60 GHz x 8). The proposed algorithm may be used in other scenarios and case studies, improving the feasibility of large-scale activity-based modeling rooted in behavioral science, thus fostering the number of similar studies/tools. The database, software, and the resulting behavioral parameters are available as open-source\textsuperscript{1}. 
Furthermore, the applicability of the proposed algorithm is not limited to activity-based modeling and transportation problems, greatly increasing its applicability.

Finally, it is still worth noting that the study has some limitations that may be addressed in future research directions. Since the calibration has been carried out against aggregate benchmarks (e.g., the modal share of the whole population), future developments may strive to apply the calibration algorithm to a more disaggregate set of data (calibrating, for example, modal share for the type of tour or type of individual) reducing local discrepancies in the modal share. Moreover, the empirically quantified uncertainty, i.e., standard deviation, of the Random forests base predictions tends to collapse to 0 in the regions of the space that are distant from the observed points. This implies similar predictions by all base models and hence inaccurately estimated uncertainty. Such phenomena may exhibit greater visibility at the initial stage of the BO, when the space is very scarcely sampled, thus future works may allow limiting the exploration to the distant regions of the parameter space.

\section*{Biography Section}
\vspace{-8mm}
\begin{IEEEbiography}
[{\includegraphics[width=1in,height=1.25in,clip,keepaspectratio]{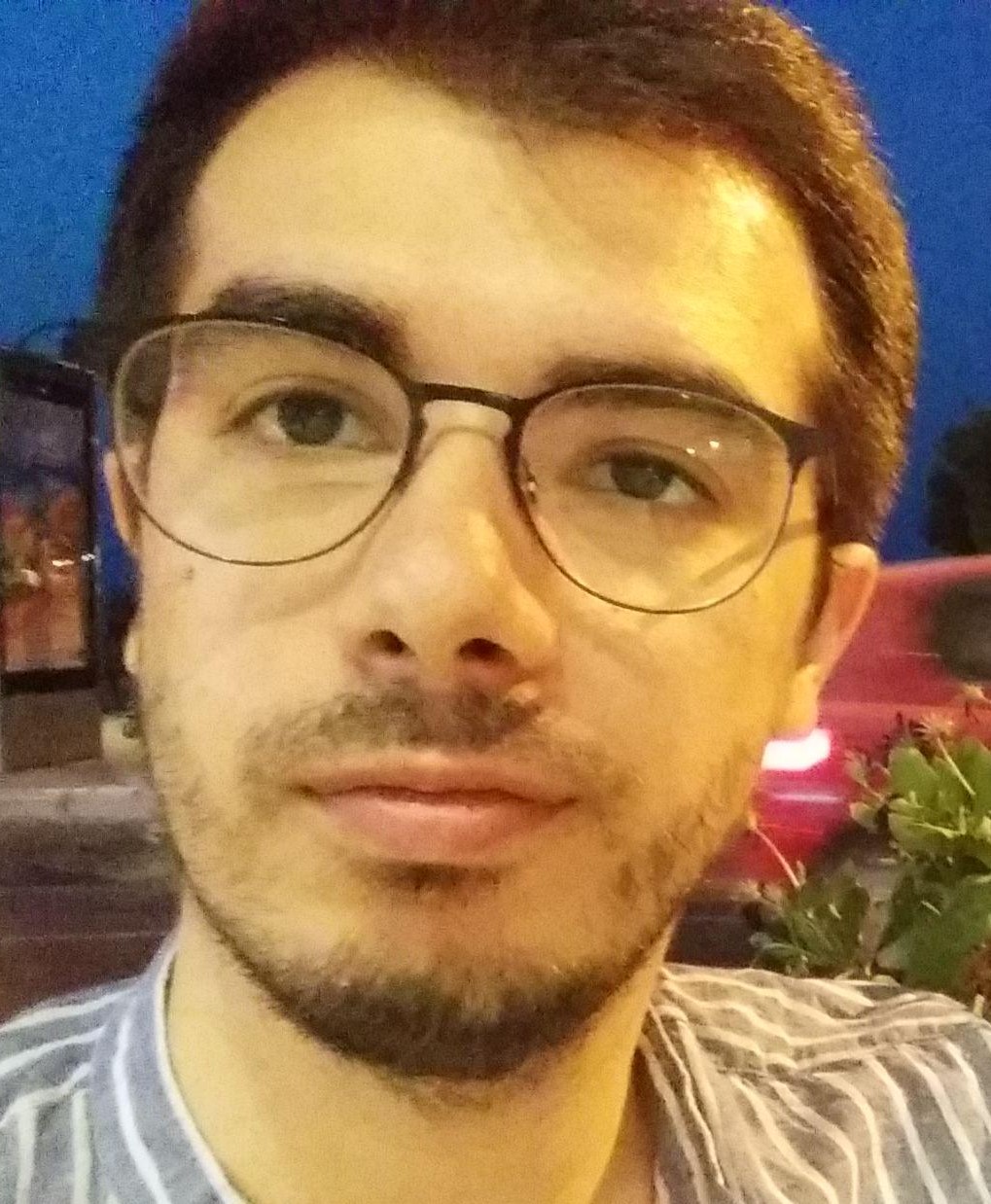}}]{Serio Agriesti}
is a PhD student at Aalto University. Before starting his doctoral studies, he was a research fellow at Politecnico di Milano, where he focused on the impact assessment of innovative transport systems such as connected and automated vehicles (CAVS) and truck platooning. He has been involved in multiple European research projects and is part of the EIT Urban Mobility Doctoral Training Network. His current research activities focus on agent-based modeling, performance evaluation and connected and automated driving.
\vspace{-3mm}
\end{IEEEbiography}

\begin{IEEEbiography}[{\includegraphics[width=1in,height=1.25in,clip,keepaspectratio]{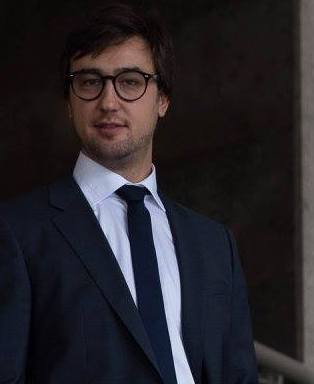}}]{Vladimir Kuzmanovski}
is a postdoctoral researcher in machine learning and artificial intelligence at Aalto University, Finland and a visiting research assistant at the Jo\v{z}ef Stefan Institute, Slovenia. He completed his PhD degree in 2016 at the International School Jo\v{z}ef Stefan, Slovenia.
His research interests include global optimization in high-dimensional spaces, surrogate modeling, probabilistic methods, and modelling of complex systems.
\vspace{-3mm}
\end{IEEEbiography}

\begin{IEEEbiography}
[{\includegraphics[trim={5cm 1cm 11cm 1cm}, width=1in,height=1.25in,clip,keepaspectratio]{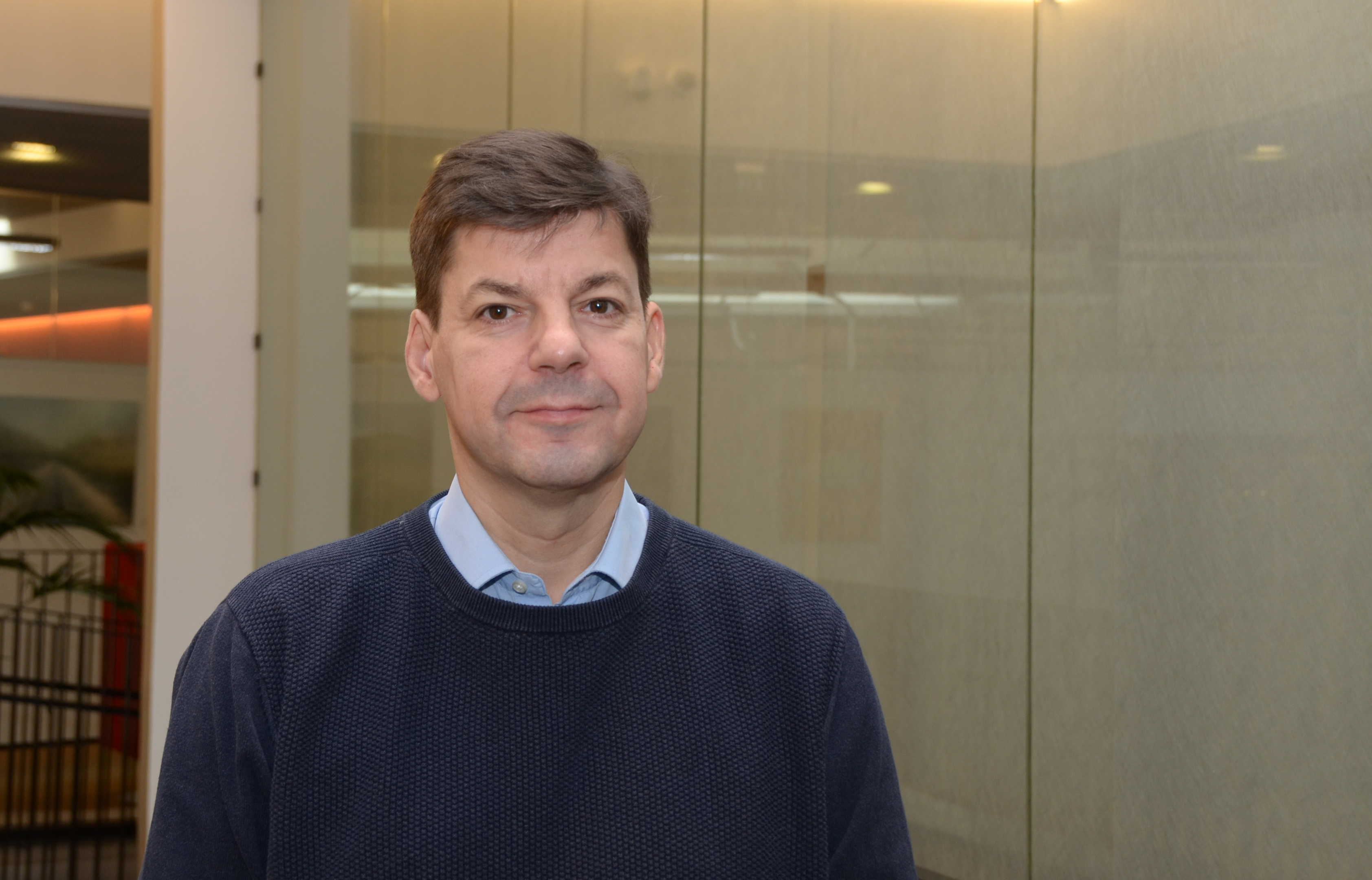}}]
{Jaakko Hollmén} is a Senior Lecturer at the Department of Computer and Systems Sciences at Stockholm University in Sweden, and Senior University Lecturer at the Department of Computer Science at Aalto University in Finland. His research interests include machine learning and data mining, and applications within health and medicine as well as environmental informatics. He is a Senior Member of the IEEE.
\vspace{-3mm}
\end{IEEEbiography}

\begin{IEEEbiography}[{\includegraphics[width=1in,height=1.25in,clip,keepaspectratio]{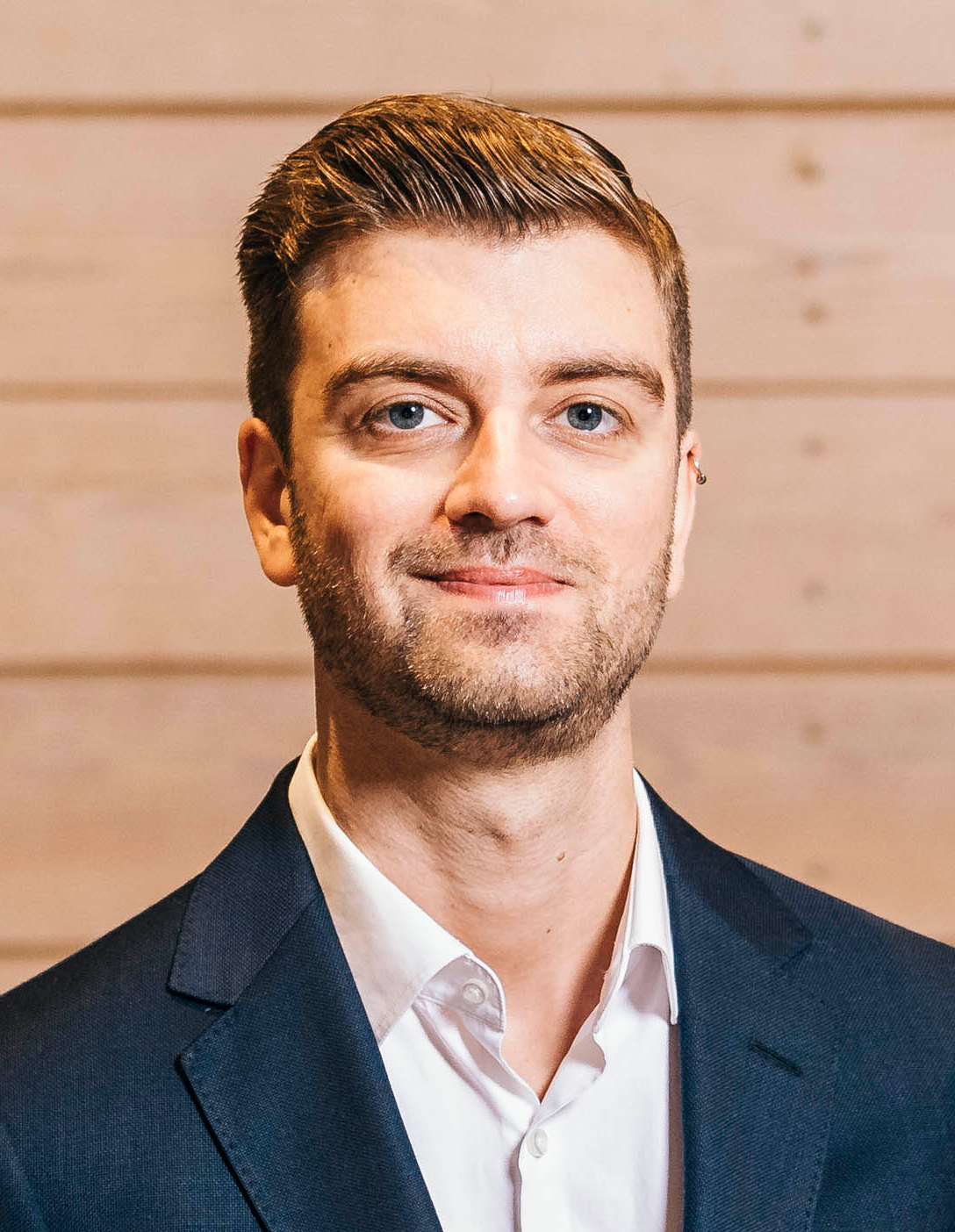}}]{Claudio Roncoli}
is an Assistant Professor of Transportation Engineering at Aalto University, Finland. He completed his PhD degree in 2013 at the University of Genova, Italy.
Before joining Aalto University, he was a research assistant at the University of Genova, Italy, a visiting research assistant at Imperial College London, UK, and a Postdoctoral Researcher at the Technical University of Crete, Greece.
Claudio has been involved in several national and international research projects as a principal investigator. His research interests include real-time traffic management; modelling, optimisation, and control of traffic systems with connected and automated vehicles; as well as smart mobility and intelligent transportation systems.
\vspace{-3mm}
\end{IEEEbiography}

\begin{IEEEbiography}[{\includegraphics[width=1in,height=1.25in,clip,keepaspectratio]{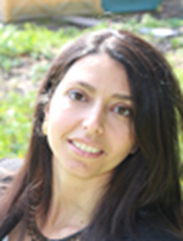}}]{Bat-hen Nahmias-Biran} is a senior lecturer in Transportation Engineering at Ariel University, Israel, and head of the Future Mobility Lab. She is also a Research Affiliate at the Intelligent Transportation Systems Lab, Massachusetts Institute of Technology (MIT). Prior to joining Ariel University, she was a Postdoctoral Associate at MIT at SMART, Future Urban Mobility (FM) lab. She has a PhD (2016) in Transportation Systems, an MSc (2011) in Transportation Engineering, and a BSc in civil and environmental Engineering (2008), all from Technion. Her main research interests are the modelling and evaluation of new technologies in transportation, simulation of shared and automated mobility, activity-based modelling, machine learning capabilities for transportation, and transport equity
\end{IEEEbiography}

\vfill

\end{document}